\documentclass{article}
\usepackage[utf8]{inputenc} % allow utf-8 input
\usepackage[T1]{fontenc}    % use 8-bit T1 fonts
\usepackage{hyperref}       % hyperlinks
\usepackage{url}            % simple URL typesetting
\usepackage{booktabs}       % professional-quality tables
\usepackage{amsfonts}       % blackboard math symbols
\usepackage{nicefrac}       % compact symbols for 1/2, etc.
\usepackage{microtype}      % microtypography
\usepackage{hyperref}
\usepackage{url}
\usepackage{times}
\usepackage{amsmath,bbm,amssymb,mathtools,amsthm}
\usepackage{algpseudocode}
\usepackage{breqn}
\usepackage{algorithm}
\usepackage{graphicx}
\usepackage{tikz, color}
\usetikzlibrary{matrix}
\usetikzlibrary{positioning}
\usetikzlibrary{graphs}
\usepackage{tabularx, environ}
\usepackage[top=90pt,bottom=90pt,left=88pt,right=88pt]{geometry}

\usepackage{amsmath,amsfonts,bm,amssymb}

\def\figref#1{fig.~\ref{#1}}
\def\Figref#1{Fig.~\ref{#1}}
\def\secref#1{sec.~\ref{#1}}
\def\eqref#1{eqn.~\ref{#1}}
\def\algref#1{algorithm~\ref{#1}}
\def\tabref#1{tab.~\ref{#1}}
\let\emptyset\varnothing

\newcommand{\E}{\mathbb{E}}

\newcommand{\softmax}{\mathrm{softmax}}

\newcommand{\KL}{D_{\mathrm{KL}}}

\DeclareMathOperator*{\argmax}{arg\,max}
\DeclareMathOperator*{\argmin}{arg\,min}

\newcommand{\mdp}{\mathcal{M}}
\newcommand{\actsp}{\mathcal{A}}
\newcommand{\ent}{\pi}
\newcommand{\ts}{\textsc{TreeSample}}

\newcommand{\scope}{\operatorname{scope}}

\newcommand{\tree}{\mathcal{T}}

\newcommand{\ie}{i.e.~}
\newcommand{\eg}{e.g.~}

\newtheorem{definition}{Definition}

\newtheorem{lemma}{Lemma}

\newtheorem{corollary}{Corollary}

\newtheorem{observation}{Observation}

\makeatletter
\newcommand{\probleminput}[1]{\gdef\@probleminput{#1}}% Store problem input
\newcommand{\problemquestion}[1]{\gdef\@problemquestion{#1}}% Store problem question
\NewEnviron{problem}{
  \probleminput{}\problemquestion{}% Default input is empty
  \BODY% Parse input
  \par\addvspace{.5\baselineskip}
  \noindent
  \begin{tabularx}{\textwidth}{@{\hspace{\parindent}} l X c}
    \textbf{Input:} & \@probleminput \\% Input
    \textbf{Output:} & \@problemquestion% Question
  \end{tabularx}
  \par\addvspace{.5\baselineskip}
}

\title{Approximate Inference in Discrete Distributions with\\ Monte Carlo Tree Search and Value Functions}
\date{}

\author{%
  Lars Buesing, Nicolas Heess, Theophane Weber\\
  DeepMind \\
  \{\texttt{lbuesing,heess,theophane}\}\texttt{@google.com}
}

\begin{document}

\maketitle

\begin{abstract}
A plethora of problems in AI, engineering and the sciences are naturally formalized as inference in discrete probabilistic models.
Exact inference is often prohibitively expensive, as it may require evaluating the (unnormalized) target density on its entire domain.
Here we consider the setting where only a limited budget of calls to the unnormalized density oracle is available, raising the challenge of where in the domain to allocate these function calls in order to construct a good approximate solution.
We formulate this problem as an instance of \emph{sequential decision-making under uncertainty} and leverage methods from reinforcement learning for probabilistic inference with budget constraints.
In particular, we propose the \ts\ algorithm, an adaptation of Monte Carlo Tree Search to approximate inference.
This algorithm caches all previous queries to the density oracle in an explicit search tree, and dynamically allocates new queries based on a "best-first" heuristic for exploration, using existing upper confidence bound methods.
Our non-parametric inference method can be effectively combined with neural networks that compile approximate conditionals of the target, which are then used to guide the inference search and enable generalization across multiple target distributions.
We show empirically that \ts\ outperforms standard approximate inference methods on synthetic factor graphs.
\end{abstract}

\section{Introduction}

Probabilistic (Bayesian) inference formalizes reasoning under uncertainty based
on first principles \cite{cox1946probability, jaynes2003probability}, with a
wide range of applications in cryptography \cite{turing1941applications},
error-correcting codes \cite{mceliece1998turbo}, bio-statistics
\cite{robinson2010edger}, particle physics \cite{baydin2019etalumis}, generative
modelling \cite{kingma2013auto}, causal reasoning \cite{pearl2000causality} and
countless others. Inference problems are often easy to formulate, \eg by
multiplying non-negative functions that each reflect independent pieces of
information, yielding an \emph{unnormalized target density} (UTD).  However,
extracting, \ie inferring, knowledge from this UTD representation, such as
marginal distributions of variables, is notoriously difficult and essentially
amounts to solving the \textsc{SumProd} problem \cite{dechter1999bucket}:
\begin{equation*} \sum_{x_1}\cdots\sum_{x_N}\prod_{m=1}^M \exp \
\psi_m(x_1,\ldots,x_N),
\end{equation*} where the UTD here is given by $\prod_m \exp \ \psi_m$.  For
discrete distributions, inference is \#P-complete \cite{roth1996hardness}, and
thus at least as hard as (and suspected to be much harder than) NP-complete
problems \cite{stockmeyer1985approximation}.

The hardness of exact inference, which often prevents its application in
practice, has led to the development of numerous approximate methods, such as
Markov Chain Monte Carlo (MCMC) \cite{hastings1970monte}, Sequential Monte Carlo
(SMC) methods \cite{del2006sequential} and Variational Inference (VI)
\cite{jordan1999introduction}.  Whereas exact inference methods essentially need
to evaluate and sum the UTD over its entire domain in the worst case,
approximate methods attempt to reduce computation by concentrating evaluations
of the UTD on regions of the domain that contribute most to the probability
mass.  The exact locations of high-probability regions are, however, often
unknown a-priori, and different approaches use a variety of means to identify
them efficiently.  In continuous domains, Hamiltonian Monte Carlo and Langevin
sampling, for instance, guide a set of particles towards high density
regions by using gradients of the target density \cite{neal2011mcmc,
roberts2002langevin}.  In addition to a-priori knowledge
about the target density (such as a gradient oracle), adaptive approximation
methods use the outcome of previous evaluations of the UTD to dynamically
allocate subsequent evaluations on promising parts of the domain
\cite{mansinghka2009exact, andrieu2008tutorial}.  This can be formalized as an
instance of \emph{decision-making under uncertainty}, where acting corresponds
to evaluating the UTD and the goal is to discover probability mass in the domain
\cite{lu2018exploration}.  Form this viewpoint, approximate inference methods
attempt to \emph{explore} the target domain based on a-priori information about
the target density as well as on \emph{partial feedback} from previous
evaluations of the UTD.

In this work, we propose a new approximate inference method for discrete
distributions, termed \ts, that is motivated by the correspondence between
probabilistic inference and decision-making highlighted previously in
the literature, \eg \cite{dayan1997using, rawlik2013stochastic,
weber2015reinforced,wingate2013automated,schulman2015gradient,weber2019credit}.
\ts\ approximates
a joint distribution over multiple discrete variables by the following
\emph{sequential decision-making} approach: Variables are inferred / sampled one variable at a
time based on all previous ones in an arbitrary, pre-specified ordering.  An
explicit tree-structured cache of all previous UTD evaluations is maintained,
and a heuristic inspired by Upper Confidence Bounds on Trees (UTC)
\cite{kocsis2006bandit} for trading off exploration around configurations that
were previously found to yield high values of UTD and configurations in regions
that have not yet been explored, is applied.  Algorithmically, \ts\ amounts to a
variant of Monte Carlo Tree Search (MCTS) \cite{browne2012survey}, modified so
that it performs integration rather than optimization.  In contrast to other
approximate methods, it leverages systematic, backtracking tree search with a
"best-first" exploration heuristic.

Inspired by prior work on combining MCTS with function approximation
\cite{silver2016mastering}, we proceed to augment \ts\ with neural networks that
parametrically cache previously computed approximate solutions of inference
sub-problems.  These networks represent approximate conditional densities and
correspond to state-action value function in decision-making and reinforcement
learning.  This caching mechanism (under suitable assumptions) allows to
\emph{generalize} search knowledge across branches of the search tree for a
given target density as well as across inference problems for different target
densities.  In particular, we experimentally show that suitably structured
neural networks such as Graph Neural Networks \cite{battaglia2018relational}
can efficiently guide the search even on new problem instances, therefore
reducing the effective search space massively.

The paper is structured as follows.  In \secref{sec:prob_notation} we introduce
notation and set up the basic inference problem.  In \secref{sec:alg}, this
inference problem is cast into the language of sequential decision-making and
the \ts\ algorithm is proposed.  We show in \secref{sec:exp} empirically, that
\ts\ outperforms closely related standard approximate inference algorithms.  We
conclude with a discussion of related work in \secref{sec:rel_work}.

\section{Discrete Inference with Computational Budget Constraints}\label{sec:prob_notation}
\subsection{Notation}
Let $X=(X_1,\ldots,X_N)\sim P^\ast_X$ be a discrete random vector taking values  $x=(x_1,\ldots,x_N)$ in $\mathbb X:=\lbrace 1,\ldots,K \rbrace^N$, and let $x_{\leq n}:=(x_1,\dots,x_n)\in \mathbb X_{\leq n}$ be its $n$-prefix and define $x_{<n}\in \mathbb X_{<n}$ analogously.
We assume the distribution $P^\ast_X$ is given by a factor graph.
Denote with $\gamma^\ast$ its density (probability mass function) and with $\hat \gamma$ the corresponding unnormalized density:
\begin{eqnarray}\label{eq:target_def}
    \log\gamma^\ast (x) = \sum_{m=1}^M \psi_m(x) - \log\sum_{x\in\mathbb X}\exp\sum_{m=1}^M \psi_m(x)=  \log\hat\gamma(x)-\log Z,
\end{eqnarray}
where $Z$ is the normalization constant.
We assume that all factors $\psi_m$, defined in the log-domain, take values in $\mathbb R\cup\{-\infty\}$. 
Furthermore, $\scope(\psi_m)$ for all $m=1,\ldots,M$ are assumed known, 
where $\scope(\psi_m)\subseteq\{1,\ldots,N\}$ is the index set of the variables that $\psi_m$ takes as input.
We denote the densities of the conditionals $P^\ast_{X_{n}\vert x_{< n}}$ as $\gamma^\ast_n(x_n\vert x_{< n})$.

\subsection{Problem Setting and Motivation}\label{sec:problem}
Consider the problem of constructing a tractable approximation $P_X$ to $P^\ast_X$.
In this context, we define tractable as being able to sample from $P_X$ (say in polynomial time in $N$).
Such a $P_X$ then allows Monte Carlo estimates of $\mathbb E_{P^\ast_X}[f]\approx\mathbb E_{P_X}[f]$ for any function $f$ of interest in downstream tasks without having to touch the original $P^\ast_X$ again. 
This setup is an example of model compilation \cite{darwiche2002logical}.
We assume that the computational cost of inference in $P^\ast_X$ is dominated by evaluating any of the factors $\psi_m$.
Therefore, we are interested in compiling a good approximation $P_X$ using a fixed computational budget:
\begin{problem}
  \probleminput{Factor oracles $\psi_1,\ldots,\psi_M$ with known $\scope(\psi_m)$; budget $B\in\mathbb N$ of pointwise evaluations of any $\psi_m$}
  \problemquestion{Approximation $P_X\approx P^\ast_X$ that allows tractable sampling}
\end{problem}
\noindent 
A brute force approach would exhaustively compute all conditionals $P^\ast_{X_1\vert \emptyset}$, $P^\ast_{X_2\vert x_1}$ up to $P^\ast_{X_N\vert x_{<N}}$ and resort to ancestral sampling.
This entails explicitly evaluating the factors $\psi_m$ everywhere, likely including "wasteful" evaluations in regions of $\mathbb X$ with low density $\gamma^\ast$, 
\ie parts of $\mathbb X$ that do not significantly contribute to $P^\ast_X$.
Instead, it may be more efficient to construct an approximation $P_X$ that concentrates computational budget on those parts of the domain $\mathbb X$ where the density $\gamma^\ast$,
or equivalently $\hat\gamma$, is suspected to be high.
For small budgets $B$, determining the points where to probe $\hat\gamma$ should ideally be done sequentially:
Having evaluated $\hat\gamma$ on values $x^1,\ldots,x^{b}$ with $b<B$, the choice of $x^{b+1}$ should be informed by the previous results $\hat\gamma(x^1),\ldots, \hat\gamma(x^{b})$.
If \eg the target density is assumed to be "smooth", a point $x$ "close" to points $x^i$ with large $\hat\gamma(x^i)$ might also have a high value $\hat\gamma(x)$ under the target, making it a good candidate for future exploration (under appropriate definitions of "smooth" and "close").
In this view, inference presents itself as a structured \emph{exploration} problem of the form studied in the literature on sequential decision-making under uncertainty and reinforcement learning,
in which we decide where to evaluate $\hat\gamma$ next in order to reduce uncertainty about its exact values.
As presented in detail in the following, borrowing from the RL literature, we will use a form of tree search that preferentially explores points $x^j$ that share a common prefix with previously found points $x^i$ with high $\hat\gamma$.

\section{Approximate Inference with Monte Carlo Tree Search}\label{sec:alg}

In the following, we cast sampling from $P^\ast_X$ as a sequential decision-making problem in a suitable maximum-entropy Markov Decision Process (MDP).
We show that the target distribution $P^\ast_X$ is equal to the solution, \ie the optimal policy, of this MDP.
This representation of $P^\ast_X$ as optimal policy allows us to leverage standard methods from RL for approximating $P^\ast_X$.
Our definition of the MDP will capture the following intuitive procedure: 
At each step $n=1,\ldots, N$ we decide how to sample $X_n$ based on the realization $x_{<n}$ of $X_{< n}$ that has already been sampled.
The reward function of the MDP will be defined such that the return (sum of rewards) of an episode will equal the unnormalized target density $\log\hat\gamma$, therefore "rewarding" samples that have high probability under the target.

\subsection{Sequential Decision-Making Representation}
We first fix an arbitrary ordering over the variables $X_1,\ldots, X_N$; for now any ordering will do, but see the discussion in \secref{sec:rel_work}. 
We then construct an episodic, maximum-entropy MDP  $\mathcal{M}=((\mathbb X_1,\ldots, \mathbb X_{\leq N}), \actsp, \circ, R^\ent)$ consisting of episodes of length $N$.
The state space at time step $n$ is $\mathbb X_{\leq n}$ and the action space is $\actsp=\{ 1,\ldots, K \}$ for all $n$.
State transitions from $x_{<n}$ to $x_{\leq n}$ are deterministic: 
Executing action $a\in\mathcal A$ in state $x_{<n}\in\mathbb X_{<n}$ at step $n$ results in setting $x_n$ to $a$, or equivalently
the action $a$ is appended to the current state, \ie $x_{\leq n} = x_{<n}\circ a=(x_1,\ldots, x_{n-1},a)$.
A stochastic policy $\pi$ in this MDP is defined by probability densities $\pi_n(a\vert x_{<n})$ over actions conditioned on $x_{<n}$ for $n=1,\ldots,N$.
It induces a joint distribution $P^\pi_X$ over $\mathbb X$ with the density $\pi(x)=\prod_{n=1}^N \pi_n(x_n\vert x_{< n})$.
Therefore, the space of stochastic policies is equivalent to the space of distributions over $\mathbb X$.

We define the maximum-entropy reward function $R^\ent$ of $\mdp$ based on the scopes of the factors $\psi_m$ as follows: 
\begin{definition}[Reward]\label{def:reward}
For $n=1\ldots,N$, we define the reward function $R_n: \mathbb X_{\leq n}\rightarrow\mathbb R\cup\{-\infty\}$, as the sum over factors $\psi_m$ 
that can be computed from $x_{\leq n}$, but not already from $x_{<n}$, \ie:
\begin{equation}\label{eq:reward_def}
    R_n(x_{\leq n}) := \sum_{\psi\in M_n}\psi(x_{\leq n}),
\end{equation}
where $M_n:=\{\ \psi_m \ \vert \ \max (\scope(\psi_m)) = n \ \}$.
We further define the maximum-entropy reward:
\begin{equation}
    R_n^\ent(x_{\leq n}) = R_n(x_{\leq n}) - \log\pi_n(x_n\vert x_{<n}).
\end{equation}
\end{definition}
To illustrate this definition, assume $\psi_1$ is only a function of $x_n$; then it will contribute to $R_n$. 
If, however, it is has full support $\scope(\psi_1)=\{1,\ldots,N\}$, then it will contribute to $R_N$. 
Evaluating $R_n$ at any input incurs a cost of $\vert M_n\vert$ towards the budget $B$.
This completes the definition of $\mdp$.
From the reward definition follows that we can write the logarithm of the unnormalized target density as the \emph{return}, \ie sum of rewards (without entropy terms):
\begin{equation}
    \log \hat\gamma(x) = \sum_{n=1}^N R_n(x_{\leq n}).
\end{equation}
We now establish that the MDP $\mathcal{M}$ is equivalent to the initial inference problem by using the standard definition of the value $V_n^\pi(x_{<n})$ of a policy as expected return conditioned on $x_{<n}$, \ie
$V_n^\pi(x_{<n}):=
  \E_{\pi}[\sum_{n'=n}^NR^\ent_{n'}]$ 
where the expectation $\E_\pi$ is taken over $P^\pi_{X_{\geq n}\vert x_{<n}}$
The following straight-forward observation holds:
\begin{observation}[Equivalence of inference and max-ent MDP]\label{obs:value}
The value $V^\pi:=V^\pi(\emptyset)$ of the initial state $x_0:=\emptyset$ under $\pi$ in the maximum-entropy MDP $\mdp$ is given by the negative KL-divergence between $P^\pi_X$ and the target $P^\ast_X$ up to the normalization constant $Z$:
\begin{equation}
    V^\pi = -\KL[P^\pi_X \Vert P^\ast_X]+\log Z.
\end{equation}
The optimal policy $\pi^\ast =\argmax_\pi V^\pi$ is equal to the target conditionals $\gamma^\ast_n(x_n\vert x_{<n})$:
\begin{eqnarray*}
    \pi_n^\ast(x_n\vert x_{<n}) &=& \gamma^\ast_n (x_n\vert x_{< n})\\
    V^{\ast} &=& \log Z.
\end{eqnarray*}
\end{observation}
\noindent Therefore, solving the maximum-entropy MDP $\mdp$ is equal to finding all target conditionals $P^\ast_{X_n\vert x_{<n}}$, and running the optimal policy $\pi^\ast$ yields samples from $P^\ast_X$.
In order to convert the above MDP into a representation that facilitates finding a solution, we use the standard definition of the state-action values as $Q^\pi_n(x_n\vert x_{<n}):=R_n(x_{\leq n})+V_{n+1}^\pi(x_{\leq n})$.
This definition together with observation \ref{obs:value} directly results in (see appendix for proof):
\begin{observation}[Target conditionals as optimal state-action values]\label{obs:q-values}
The target conditional is proportional to the optimal state-action value function, \ie $\gamma^\ast_n(x_n\vert x_{<n})=\exp(Q_n^\ast(x_n\vert x_{<n})-V^\ast_n(x_{<n}))$ where the normalizer is given by the value $V^\ast_n(x_{<n})=\log\sum_{x_n}\exp Q_n^\ast(x_n\vert x_{<n})$. 
Furthermore, the optimal state-action values obey the \emph{soft Bellman equation}:
\begin{equation}\label{eq:soft-B}
    Q_n^\ast(x_n\vert x_{<n})  \ = \ R_{n}(x_{\leq n}) + \log\sum_{x_{n+1}=1}^K\exp \  Q^\ast_{n+1}(x_{n+1}\vert x_{\leq n}).
\end{equation}
\end{observation}

\subsection{\ts\ Algorithm}

In principle, the soft-Bellman equation \ref{eq:soft-B} can be solved by backwards dynamic programming in the following way.
We can represent the problem as a $K$-ary tree $\tree^\ast$ over nodes corresponding to all partial configurations $\bigcup_{n=0}^N \mathbb X_{\leq n}$, root $\emptyset$ and each node $x_{<n}$ being the parent of $K$ children $x_{<n}\circ 1$ to $x_{<n}\circ K$.
One can compute all $Q$-values by starting from all $K^N$ leafs $x_{\leq N}\in\mathbb X_N$ for which we can compute the state-action values $Q_N^\ast(x_N\vert x_{< N})=R_N(x)$ and solve \eqref{eq:soft-B} in reverse order.
Furthermore, a simple softmax operation on each $Q^\ast_n$ yields the target conditional $\gamma_n^\ast(x_n\vert x_{<n})$.
Unfortunately, this requires exhaustive evaluation of all factors.

\begin{algorithm}[t] 
  \caption{\ts\ sampling procedure}
\label{alg:tree_sample}
\begin{algorithmic}[1]
\Procedure{Sample}{tree $\tree$, default state-action values $Q^\phi$}
    \State $x\leftarrow \emptyset$
    \For{$n=1,\ldots,N$}
        \If{$x \in \tree$}
            \State $a\sim \softmax \ \  Q_n(\cdot\vert x)$
        \Else
            \State $a\sim \ \softmax \ \  Q_n^\phi(\cdot\vert x)$
        \EndIf    
        \State $x \leftarrow x \circ a$            
        \EndFor
        \State {\bf return} $x$ 
\EndProcedure
\end{algorithmic}
\end{algorithm}

As an alternative to exhaustive evaluation, we propose the \ts\ algorithm for approximate inference.
The main idea is to construct an approximation $P_X$ consisting of a \emph{partial tree} $\tree\subseteq \tree^\ast$ and approximate state-actions values $Q$ with support on $\tree$.
A node in $\tree$ at depth $n$ corresponds to a prefix $x_{< n}$, with the attached vector of state-action values 
$Q_n(\cdot\vert x_{<n})=(Q_n(x_{n}=1\vert x_{<n}), \ldots, Q_n(x_{n}=K\vert x_{<n}))\approx Q^\ast(\cdot\vert x_{<n})$ for its $K$ children $x_{<n}\circ 1$ to $x_{<n}\circ K$ (which might not be in tree themselves).
Sampling from $P_X$ is defined in \algref{alg:tree_sample}:
The tree is traversed from the root $\emptyset$ and at each node,
a child is sampled from the softmax distribution defined by $Q$.
If at any point, a node $x_{\leq n}$ is reached that is not in $\tree$, 
the algorithm falls back to a distribution defined by a user-specified, default state-action value function $Q^\phi$; 
we will also refer to $Q^\phi$ as prior state-action value function as it assigns a state-action value before / without any evaluation of the reward.
Later, we will discuss using learned, parametric functions for $Q^\phi$.
In the following we describe how the partial tree $\tree$ is constructed using a given, limited budget of $B$ of evaluations of the factors $\psi_m$.

\subsubsection{Tree Construction with Soft-Bellman MCTS}\label{sec:tree_construction}
\ts\ leverages the correspondence of approximate inference and decision-making that we have discussed above.
It consists of an MCTS-like algorithm to iteratively construct the tree $\tree$ underlying the approximation $P_X$.
Given a partially-built tree $\tree$, 
the tree is expanded (if budget is still available) using a heuristic inspired by Upper Confidence Bound (UCB) methods \cite{auer2002finite}.
It aims to expand the tree at branches expected to have large contributions to the probability mass by taking into account how important a branch currently seems, given by its current $Q$-value estimates, as well as a measure of uncertainty of this estimate. The latter is approximated by a computationally cheap heuristic based on the visit counts of the branch, \ie how many reward evaluations have been made in this branch.
The procedure prefers to explore branches with high $Q$-values and high uncertainty (low visit counts); it is given in full in \algref{alg:ts-details} in the appendix, but is briefly summarized here.

Each node $x_{<n}$ in $\tree$,  in addition to $Q_n(\cdot\vert x_{<n})$, also keeps track of its visit count $\eta(x_{<n})\in \mathbb N$ and the cached reward evaluation $R_{n-1}(x_{<n})$.
For a single tree expansion, $\tree$ is traversed from the root by choosing at each intermediate node $x_{<n}$ the next action $a\in\{1\ldots, K\}$ in the following way:
\begin{eqnarray}\label{eq:q-uct}
   \argmax_{a=1,\ldots, K}\ \ \ Q_n(a\vert x_{<n}) 
   + c\cdot\max( Q_n^\phi(a\vert x_{<n}), \epsilon )\frac{\eta(x_{<n})^{1/2}}
   {1+\eta(x_{<n}\circ a)}.
\end{eqnarray}
Here, the hyperparameters $c>0$ and $\epsilon>0$ determine the influence of the second term, which can be seen as a form of exploration bonus and which is computed from the inverse visit count of the action $a$ relative to the visit counts of the patent.
This rule is inspired by the PUCT variant employed in \cite{silver2016mastering}, 
but using the default value $Q^\phi$ for the exploration bonus.
When a new node $x^{new}\not\in\tree$ at depth $n$ is reached, 
the reward function $R_n(x^{new})$ is evaluated, decreasing our budget $B$.
The result is cached and the node is added $\tree\leftarrow \tree \cup x^{new}$ using $Q^\phi$ to initialize $Q_{n+1}(\cdot\vert x^{new})$. 
Then the $Q$-values are updated: On the path of the tree-traversal that led to $x^{new}$, the values are back-upped in reverse order using the soft-Bellman equation.
This constitutes the main difference to standard MCTS methods, which employ
max- or averaging backups. 
This reflects the difference of sampling / integration to the usual application of MCTS to maximization / optimization problems.
Once the entire budget is spent, $\tree$ with its tree-structured $Q$ is returned.

\subsubsection{Consistency}
As argued above, the exact conditionals $\gamma^\ast_{n+1}(\cdot\vert x_{\leq n})$ can be computed by exhaustive search in exponential time.
Therefore, a reasonable desideratum for any inference algorithm is that given a large enough budget $B\geq MK^N$ the exact distribution is inferred.
In the following we show that \ts\ passes this basic sanity check.
The first important property of \ts\ is that a tree $\tree$ has the exact conditional $\gamma^\ast_{n+1}(\cdot\vert x_{\leq n})$ if the unnormalized target density has been evaluated on all states with prefix $x_{\leq n}$ during tree construction.
To make this statement precise, we define $\tree_{x_\leq n}\subseteq \tree$ as the sub-tree of $\tree$ consisting of node $x_{\leq n}$ and all its descendants in $\tree$. 
We call a sub-tree $\tree_{x_\leq n}$ fully expanded, or complete, if all partial states with prefix $x_{\leq n}$ are in  $\tree_{x_\leq n}$.
With this definition, we have the following lemma (proof in the appendix):
\begin{lemma}\label{lem:exact_values}
Let $\tree_{x_{\leq n}}$ be a fully expanded sub-tree of $\tree$.
Then, for all nodes $x'_{\leq m}$ in $\tree_{x_{\leq n}}$, \ie $m\geq n$ and $x'_{\leq n}=x_{\leq n}$, the state-action values are exact and in particular the node $x_{\leq n}$ has the correct value:
\begin{eqnarray*}
     Q_{m+1}(\cdot\vert x'_{\leq m}) &=&  Q_{m+1}^\ast(\cdot\vert x'_{\leq m})\\
     V_{n+1}(x_{\leq n}) &=&  V_{n+1}^\ast(x_{\leq n}) 
\end{eqnarray*}
\end{lemma}
Furthermore, constructing the full tree $\tree^\ast$ with \ts\ incurs a cost of at most $MK^N$ evaluations of any of the factors $\psi_m$,
as there are $K^N$ leaf node in $\tree^\ast$ and constructing the path from the root  $\emptyset$ to each leaf requires at most $M$ oracle evaluations.
Therefore, \ts\ with $B\geq MK^N$ expands the entire tree and the following result holds:
\begin{corollary}[Exhaustive budget consistency]
\ts\ outputs the correct target distribution $P^\ast_X$ for budgets $B\geq MK^N$ .
\end{corollary}

\subsection{Augmenting \ts\ with Learned Parametric Priors}
$\ts$ explicitly allows for a "prior" $Q^\phi$ over state-action values with parameters $\phi$.
It functions as a parametric approximation to $Q^\ast_n\propto\log\gamma^\ast_n$.
In principle, an appropriate $Q^\phi$ can guide the search towards regions in $\mathbb X$ where probability mass is likely to be found a-priori by the following two mechanisms.
It scales the exploration bonus in the PUCT-like decision rule \eqref{eq:q-uct},
and it is used to initialize the state-action values $Q$ for a newly expanded node in the search tree. 
In the following we discuss scenarios and potential benefits of learning the parameters $\phi$.

In principle, if $Q^\phi$ comes from an appropriate function class, 
it can \emph{transfer} knowledge within the inference problem at hand.
Assume we spent some of the available search budget on \ts\ to build an approximation $\tree$. 
Due to the tree-structure, search budget spent in one branch of the tree does not benefit any other sibling branch. 
For many problems, there is however structure that would allow for \emph{generalizing} knowledge across branches.
This can be achieved via $Q^\phi$, \eg
one could  train $Q^\phi$ to approximate the $Q$-values of the current $\tree$,
and (under the right inductive bias) knowledge would transfer to newly expanded branches.
A similar argument can be made for parametric generalization across problem instances.
Assume a given a family of distributions $\{P^i_X\}_{i\in I}$ for some index-set $I$.
If the different distributions $P_X^i$ share structure, it is possible to leverage search computations performed on $P^i_X$ for inference in $P^j_X$ to some degree.
A natural example for this is posterior inference in the same underlying model conditioned on different evidence / observations, similar e.g.\ to amortized inference in variational auto-encoders \cite{kingma2013auto}.
Besides transfer, there is a purely computational reason for learning a parametric $Q^\phi$.
The memory footprint of $\ts$ grows linearly with the search  budget $B$. 
For large problems with large budgets $B\gg 0$, storing the entire search tree in memory might not be feasible.
In this case, compiling the current tree periodically into $Q^\phi$ and 
rebuilding it from scratch under prior $Q^\phi$ and subsequent refinement using \ts\ may be preferable.

Concretely, we propose to train $Q^\phi$ by regression on state-action values $Q$ generated by \ts.
For generalization across branches, $Q$ approximates directly the distribution of interest, 
for transfer across distributions, $Q$ approximates the source distribution,
and we apply the trained $Q^\phi$ for inference search in a different target distribution. 
We match $Q^\phi$ to $Q$ by minimizing the expected difference of the values:
\begin{eqnarray*}
    \phi^\ast = &\argmin_{\phi} & \mathbb E_{P_X}
    \left [\sum_{n=1}^N\Vert Q_n^\phi(\cdot\vert X_{<n})-Q_n(\cdot\vert X_{<n})\Vert_2^2\right].
\end{eqnarray*}
In practice we optimize this loss by stochastic gradient descent in a distributed learner-worker architecture detailed in the experimental section.

\section{Experiments}\label{sec:exp}
\begin{figure}
    \centering
    \includegraphics[width=0.49\textwidth]{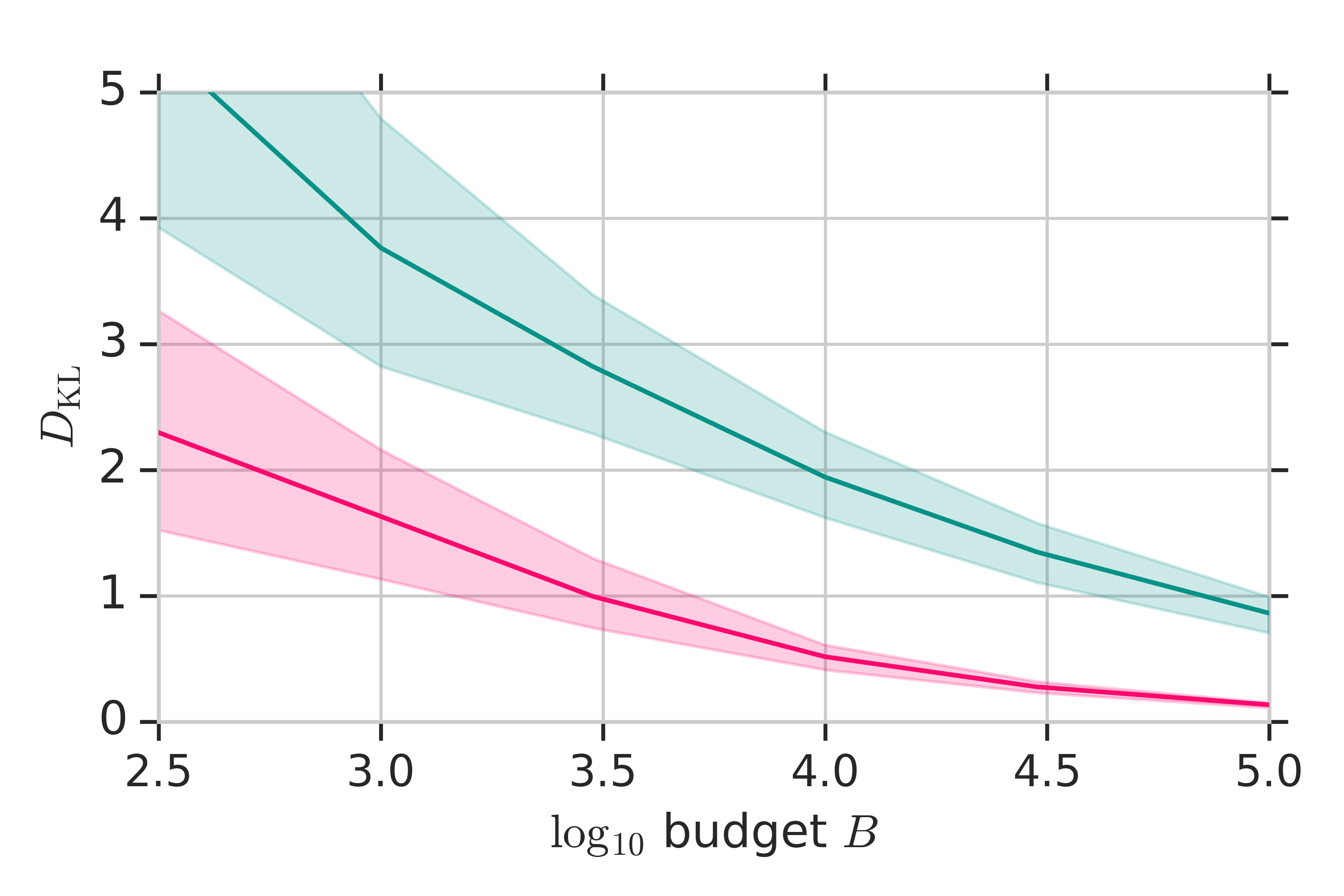}
    \includegraphics[width=0.49\textwidth]{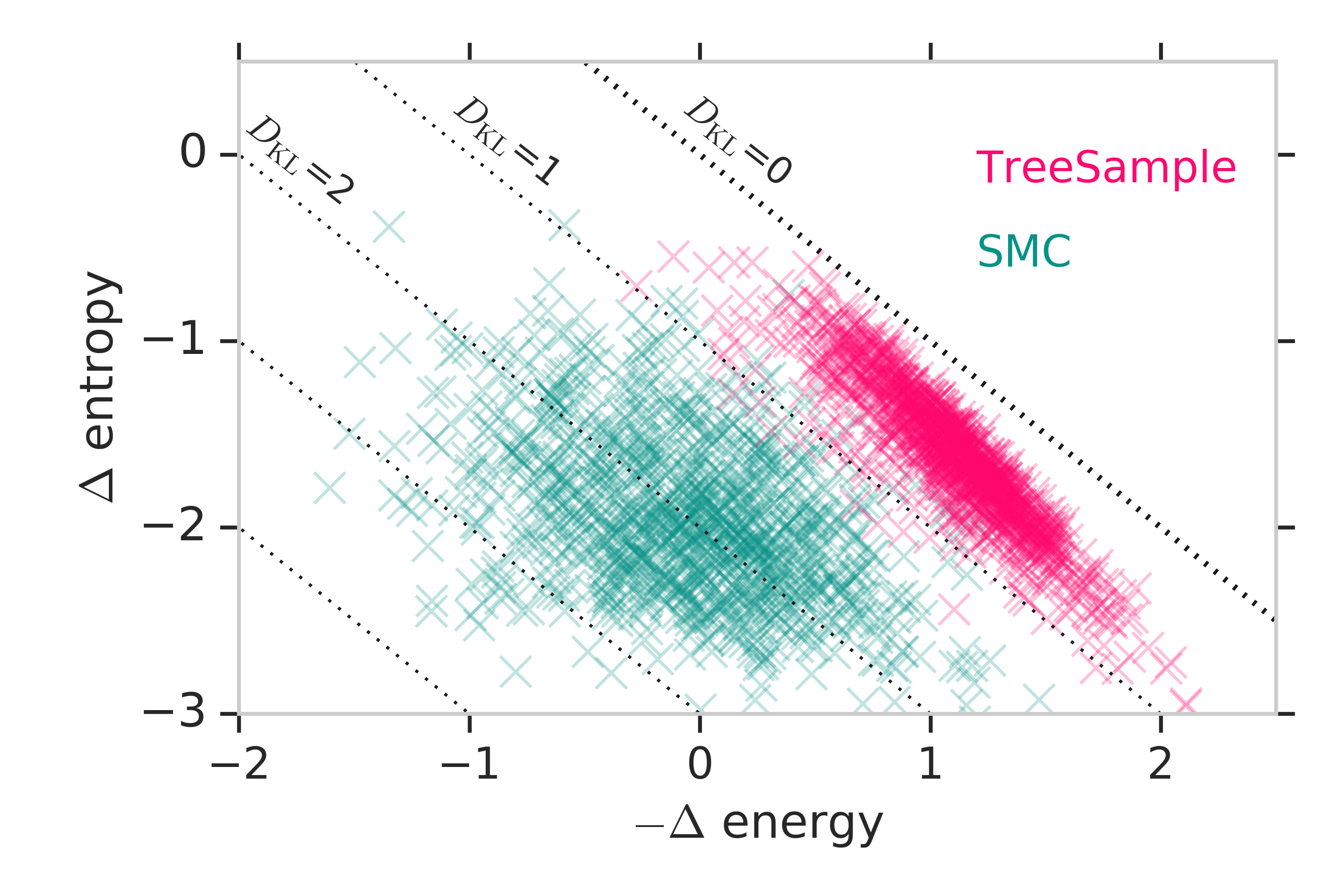}
    \caption{
        Comparison of \ts\ to SMC on inference in 1000 randomly generated Markov chains. 
        Left: Approximation error $\KL[P_X\Vert P^\ast_X]$ as a function of the budget $B$ in log-scale, showing that SCM needs more than 30 times the budget of \ts\ to generate comparable approximations.
        Right: Energy and entropy contributions to the $\KL[P_X\Vert P^\ast_X]$ for all 1000 experiments for $B=10^4$, showing that \ts\ finds approximations with both higher entropy and lower energy.
    }
    \label{fig:chain_results}
\end{figure}

In the following, we empirically compare \ts\ to other baseline inference methods on different families of distributions. 
We quantify approximation error by the Kullback-Leibler divergence:
\begin{eqnarray}\label{eq:dkl_decomp}
    \KL[P_X\Vert P^\ast_X] = & 
    \log Z&\underbrace{-\mathbb E_{P_X}\left[\sum_{m=1}^M\psi_m(X)\right]
    -\mathbb H\left[P_X\right]}_{:=}\\
    =& \log Z& + \ \ \ \ \ \ \ \Delta\KL[P_X\Vert P^\ast_X],\nonumber
\end{eqnarray}
where we refer to the second term in \eqref{eq:dkl_decomp} as negative expected energy, and the last term is the entropy of the approximation.
We can get unbiased estimates of these using samples from $P_X$.
For intractable target distributions, we compare different inference methods
using $\Delta\KL:=\KL-\log Z$, which is tractable to approximate and preserves ranking of different approximation methods.

As baselines we consider the following: 
Sequential Importance Sampling (SIS), Sequential Monte Carlo (SMC) and for a subset of the 
environments also Gibbs sampling (GIBBS) and sampling with loopy belief propagation (BP); details are given in the appendix.
We use the baseline methods in the following way:
We generate a set of particles $\{x^i\}_{i\leq I}$ of size $I$ such that we exhaust the budget $B$, and then return the (potentially weighted) sum of atoms $\sum_{i\leq I} p^i\delta(x,x^i)$ as the approximation density; 
here $\delta$ is the Kronecker delta, and $p^i$ are either set to $1/I$ for GIBBS, BP and to the self-normalized importance weights for SIS and SMC.
Hyperparameters for all methods where tuned individually for different families of distributions on an initial set of experiments and then kept constant across all reported experiments. 
For further details, see the appendix.
For SIS and SMC, the proposal distribution plays a comparable role to the state-action prior in \ts.
Therefore, for all experiments we used the same parametric family for $Q^\phi$ for \ts, SIS and SMC.

For the sake of simplicity, in the experiments we measured and constrained the inference budget $B$ in terms of reward evaluations,
\ie each pointwise evaluate of a $R_n$ incurs a cost of one, instead of factor evaluations.

\subsection{\ts\  without Parametric Value Function}

We first investigated inference without learned parametric $Q^\phi$.
Instead, we used the simple heuristic of setting $\forall a\forall n \ Q^\phi_{n}(a\vert x_{<n}):=(N-n)\log K$,
which corresponds to the state-action values when all factors $\psi_m\equiv 0$ vanish everywhere.

\subsubsection{Chain Distributions}

We initially tested the algorithms on inference in chain-structured factor
graphs (\textsc{CHAINS}). 
These allow for exact inference in linear time, and therefore we can get unbiased estimates of the true Kullback-Leibler divergences. 
We report results averaged over $10^3$ different chains of length $N=10$ with randomly generated unary and binary potential functions;
for details, see appendix.
The number of states per variable was set to $K=5$, yielding $K^N\approx 10^7$ states in total.
The results, shown in \figref{fig:chain_results}  as a function of the inference budget $B$,
show that \ts\ outperforms the SMC baseline (see also \tabref{tab:chain_results}).
In particular,  \ts\ generates approximations of similar quality compared to SMC with a roughly 30 times smaller budget.
We further investigated the energy and entropy contributions to $\KL$ separately.
We define $\Delta$energy$=\mathbb E_{P^\ast_X}[\sum\psi]-\mathbb E_{P_X}[\sum\psi]$ (lower is better), 
and $\Delta$entropy$=\mathbb H[P_X]-\mathbb H[P^\ast_X]$ (higher is better). 
\Figref{fig:chain_results} shows that \ts\ finds approximations that have lower energy as well as higher entropy compared to SMC.

\begin{table}[t]
    \centering
    \begin{tabular}{c||c|c|c|c}
        $\KL$ or $\Delta\KL$   & CHAIN                 & PERMUTED CHAIN            & FACTOR GRAPHS 1         & FACTOR GRAPHS 2   \\ \hline\hline
        SIS           & 11.61 $\pm$ 1.74      & 9.23 $\pm$ 0.34           & -21.97 $\pm$ 2.47       & -31.70 $\pm$ 2.32 \\
        SMC           & 1.94 $\pm$ 0.48       & 7.08 $\pm$ 0.36           & -24.09 $\pm$ 2.85       & -35.90 $\pm$ 2.47 \\
        GIBBS         &   --                  &        --                 & -18.67 $\pm$ 1.80       & -25.12 $\pm$ 1.48 \\
        BP            & {\small exact}        & {\small exact}            & -21.50 $\pm$ 0.18       & -31.48 $\pm$ 0.48 \\
        \ts           & {\bf 0.53 $\pm$ 0.17} & {\bf 3.41 $\pm$ 0.41}     & {\bf -28.89 $\pm$ 1.94} & {\bf -38.70 $\pm$ 2.29}
        
    \end{tabular}
    \caption{Approximation error (lower is better) for different inference methods on four distribution classes. 
    Results are averages and standard deviations over 1000 randomly generated distributions for each class.
    Budget was set to $B=10^4$.}
    \label{tab:chain_results}
\end{table}

A known limitation of tree search methods is that they tend to under-perform for  
shallow (here small $N$) decision-making problems with large action spaces (here large $K$).
We performed experiments on chain distributions with varying $K$ and $N$ while keeping
the state-space size approximately constant, \ie $N\log K\approx \mbox{const}$.
We confirmed that for very shallow, bushy problems with  $\log K\gg N$, 
SMC outperforms \ts, whereas \ts\ dominates SMC in all other problem configurations, 
see \figref{fig:vary_depth} in the appendix.

Next, we considered chain-structured distributions where the indices of the variables $X_n$
do not correspond to the ordering in the chain; we call these \textsc{PermutedChains}. 
These are in general more difficult to solve as they exhibit "delayed" rewards,
\ie binary chain potentials $\psi_{m}(X_{\sigma(n)}, X_{\sigma(n+1)})$ depend on non-consecutive variables.
This can create "dead-end" like situations, that SMC, not having the ability to backtrack, can get easily stuck in.
Indeed, we find that SCM performs only somewhat better on this class of distributions than SIS, whereas \ts\ 
achieves better results by a wide margin.
Results on both families of distributions are shown in \tabref{tab:chain_results}.

\subsubsection{Factor Graphs}

We also tested the inference algorithms on two classes of non-chain factor graphs, denoted as \textsc{FactorGraphs1} and 
\textsc{FactorGraphs2}.
Distributions in \textsc{FactorGraphs1} are over $N=10$ variables with $K=5$ states each.
Factors were randomly generated with maximum degree $d$ of 4 and their $d^K$ values where iid drawn from $\mathcal N(0,1)$. 
Distributions in \textsc{FactorGraphs2} are over $N=20$ binary variables, \ie $K=2$.
These distributions are generated by two types of of factors: NOT (degree 2) and MAJORITY (max degree 4), both taking values in $\{0,1\}$. 

Results are shown in \tabref{tab:chain_results}.
For both families of distributions, \ts\ outperforms all considered baselines by a wide margin.
We found that GIBBS generally failed to find configurations with high energy due to slow mixing.
BP-based sampling was observed to generate samples with high energy but small entropy, 
yielding results comparable to SIS.

\subsection{\ts\ with Parametric Value Functions}

\begin{table}[t]
\begin{center}
\begin{tabular}{c||c|c|c|c|c|c|c}
    {\small value func.} &
    \multicolumn{1}{c|}{$\emptyset$}  &
    \multicolumn{1}{c|}{MLP}  &
    \multicolumn{5}{c}{GNN} \\\hline
    {\small single graph} & N/A & Yes & Yes & No & No & No & No\\
    $N_{\mathrm{train}}$ & N/A & 20 & 20 & 20 & 12 & 16 & 24 \\ \hline\hline
    $Q^\phi$ trained by SMC: $\Delta\KL^\phi$ &
    --                     &   +1.63                   &   -0.19                   &   -0.97                &   -1.00                   &   -1.17                   &   -0.64 \\ &
    --                     &   {\tiny [-1.60,+2.87]}   &   {\tiny[-2.68,+1.49]}    &   {\tiny[-2.41,+0.40]} &   {\tiny[-1.52, -0.58]}   &   {\tiny [-1.42,-0.46]}   &   {\tiny[-0.84,-0.32]}\\
    $Q^\phi$ + SMC: $\Delta\KL$  &
    +2.72                  &   +2.93                   &   +1.64                   &   +2.56                &   +2.00                   &   +1.64                   &   +2.10  \\ &
    {\tiny [1.02, 4.54]}   &   {\tiny [-1.54,+4.32]}   &   {\tiny[-1.61,+3.46]}    &   {\tiny[+0.58,+4.42]} &   {\tiny[+1.72, +2.19]}   &   {\tiny [+1.38,+2.05]}   &   {\tiny[+1.50,+2.68]}    
    \\\hline
    $Q^\phi$ trained by \ts: $\Delta\KL^\phi$  &
    --                     &   -3.61                   &   -3.86                   &   -2.05                &   -2.12                   &   -2.52                   &   -1.83 \\ &
    --                     &   {\tiny[-5.73,-0.60]}    &   {\tiny[-6.03, -0.85]}   & {\tiny[-3.58, -0.55]}  &   {\tiny[-2.23,-1.99]}    &   {\tiny[-2.63,-2.40]}    &   {\tiny[-2.13,-1.76]} \\
    $Q^\phi$ + \ts: $\Delta\KL$  &
    {\bf 0.00}             &   {\bf -3.63}             &   {\bf -3.87}             &   {\bf -2.23}          &   {\bf -2.22}             &   {\bf -2.64}             &   {\bf -2.35} \\ &
    {\tiny [-1.47, 1.68]}  &   {\tiny[-5.72,-0.64]}    &   {\tiny[-6.05, -0.88]}   & {\tiny[-3.75, -0.73]}  &   {\tiny[-2.30,-2.05]}    &   {\tiny[-2.79,-2.55]}    &   {\tiny[-2.46,-2.10]}
\end{tabular}
\end{center}
\caption{
    Approximation error for inference in factor graphs with \ts\ and SMC, for different types of value functions and training regimes. Results are relative to \ts\ w/o value function, lower is better. See main text for details.}
\label{tab:value_function}
\end{table}

Next, we investigated the performance of \ts, as well as SMC, with additional parametric state-action value functions $Q^\phi$ (used as proposal for SMC).
We focused on inference problems from \textsc{FactorGraphs2}.
We implemented the inference algorithm as a distributed  architecture consisting of a \emph{worker}  and a \emph{learner} process, both running simultaneously.
The worker requests an inference problem instance, and performs inference either with  \ts\ or SMC with a small budget of $B=2500$ using the current parametric $Q^\phi$. 
After building the approximation $P_X$, 128 independent samples $x^i\sim P_X$ are drawn from it and the inferred $Q$-values $Q_{n}(\cdot\vert x^i_{\leq n})$ for $i=1,\ldots,128$ and $n=1,\ldots,N$ are written into a replay buffer as training data; 
following this, the inference episode is terminated, the tree is flushed and a new episode starts.
The learner process samples training data from the replay for updating
the parametric $Q^\phi$ with an SGD step on a minibatch of size 128; then the updated model parameters $\phi$ are sent to the worker.
We tracked the error of the inference performed on the worker using the
unnormalized $\Delta \KL$ as a function of the number of completed inference episodes.
We expect $\Delta\KL$ to decrease, as $Q^\phi$ adapts to the inference problem, and therefore becomes better at guiding the search.
Separately, we also track the inference performance of only using the value function $Q^\phi$ without additional search around it, denoted as $\Delta\KL^\phi$.
This is a purely parametric approximation to the inference problem, trained by samples from \ts\ and SMC respectively.
We observed that $\Delta\KL$ as well as $\Delta\KL^\phi$ stabilized after roughly 1500 inference episodes for all experiments. Results were then averaged over episodes 2000-4000 and are shown in \tabref{tab:value_function}.
To facilitate comparison, all results in \tabref{tab:value_function} are reported relative to $\Delta\KL$ for \ts\ without value functions.
In general, experimental results with learned value functions exhibited higher degrees of variability with some outliers. 
Results in \tabref{tab:value_function} therefore report median results over 20 runs as well as 25\% and 75\% percentiles.

We first performed a simple set of "sanity-check" experiments on \ts\ with parametric value functions in a non-transfer setting, where the worker repeatedly solves the same inference problem arising from a single factor graph.
As value function, we used a simple MLP with 4-layers and 256 hidden units each.
As shown in the second column of \tabref{tab:value_function}, 
approximation error $\Delta\KL$ decreases significantly compared to plain \ts\ without value functions. 
This corroborates that the value function can indeed cache part of the previous search trees and facilitate inference if training and testing factor graphs coincide.
Furthermore, we observed that once $Q^\phi$ is fully trained, the inference error $\Delta\KL^\phi$ obtaind using only $Q^\phi$ is only marginally worse than $\Delta\KL$ using $Q^\phi$ plus \ts-search on top of it; see row four and five in \tabref{tab:value_function} respectively.
This indicates that the value function was powerful enough in this experiment to almost cache the entire search computation of \ts.

Next, we investigated graph neural networks (GNNs) \cite{battaglia2018relational} as value functions $Q^\phi$.
This function class can make explicit use of the structure of the factor graph instances.
Details about the architecture can be found in \cite{battaglia2018relational} and the appendix, but are briefly described in the following.
GNNs consist of two types of networks, node blocks and edge blocks (we did not use global networks), that are connected according to the factor graph at hand, and executed multiple times mimicking a message-passing like procedure.
We used three node block networks, one for each type of graph node, \ie variable node (corresponding to a variable $X_n$), NOT-factors and MAJORITY-factors.
We used four edge block networks, namely one for each combination of \{incoming,outgoing\}$\times$\{NOT, MAJORITY\}.
Empirically, we found that GNNs slightly outperform MLPs in the non-transfer setting, see third column of  \tabref{tab:value_function}.

The real advantage of GNNs comes into play in a transfer setting, when the worker performs inference in a new factor graph for each episode. We keep the number of variables fixed ($N_{\mathrm{train}}=N_{\mathrm{test}}=20$) but vary the number and configuration of factors across problems.
GNNs successfully generalize across graphs, see fourth column of \tabref{tab:value_function}.
This is due to their ability to make use of the graph topology of a new factor graph instance, by connecting its constituent node and edge networks accordingly.
Furthermore, the node and edge networks evidently learned generic message passing computations for variable nodes as well as NOT/MAJORITY factor nodes.
The results show that a suitable $Q^\phi$ generalizes knowledge across inference problems, leading to less approximation error on new distributions.
Furthermore, we investigated a transfer setting where the worker solves inference problems on factor graphs of sizes $N_{\mathrm{train}}=12,16$ or 24, but performance is tested on graphs of size $N_{\mathrm{test}}=20$; see columns five to seven in \tabref{tab:value_function}.
Strikingly, we find that the value functions generalize as well across problems of different sizes as they generalize across problems of the same size.
This demonstrates that prior knowledge can successfully guide the search and greatly facilitate inference.

Finally, we investigated the performance of SMC with trained value functions $Q^\phi$; see rows one and two in \tabref{tab:value_function}.
Overall, we found that performance was worse compared to \ts: 
Value functions $Q^\phi$ trained by SMC were found to give worse results $\Delta\KL^\phi$ compared to those trained by \ts, and overall inference error was worse compared to \ts.
Interestingly, we found that once $Q^\phi$ is fully trained, performing additional SMC on top of it made results worse.
Although initially counter-intuitive, these results are sensible in our problem setup.
The entropy of SMC approximations $\approx \log I$ is essentially given by the number of particles $I$ that SMC produces; this number is limited by the budget $B$ that can be used to compute importance weights.
Once a parametric $Q^\phi$ is trained, it does not need to make any further calls to the $\psi_m$ factors, and can therefore exhibit much higher entropy, therefore making $\Delta\KL^\phi$ smaller than $\Delta\KL$.

\section{Related Work}\label{sec:rel_work}

\ts\ is based on the connection between probabilistic inference and maximum-entropy decision-making problems established by previous work.
This connection has mostly been used to solve RL problems with inference methods \eg \cite{dayan1997using, attias2003planning, hoffman2007trans, rawlik2013stochastic}.
Closely related to our approach, this relationship has also been used in the reverse direction, \ie to solve inference problems using tools from RL \cite{mnih2014neural, weber2015reinforced,wingate2013automated,schulman2015gradient,weber2019credit}, however without utilizing tree search and emphasizing the importance of exploration for inference.
The latter has been recognized in \cite{lu2018exploration}, and applied to hierarchical partitioning for inference in continuous spaces, see also \cite{rainforth2018inference}.
In contrast to this, we focus on discrete domains with sequential decision-making utilizing MCTS and value functions.
Soft-Bellman backups, as used here (also referred to as soft Q-learning) and their connection to entropy-regularized RL have been explored in \eg \cite{schulman2017equivalence, haarnoja2018soft}.

For approximating general probabilistic inference problems, the class of Markov Chain Monte Carlo (MCMC) methods has proven very successful in practice.
There, a transition operator is defined such that the target distribution is stationary under this operator.
Concretely, MCMCs methods operate on a fully specified, approximate sample which is then perturbed iteratively. 
Transition operators are usually designed specifically for families of distributions in order to leverage problem structure for achieving fast mixing.
However, mixing times are difficult to analyze theoretically and hard to monitor in practice \cite{cowles1996markov}. 
\ts\ circumvents the mixing problem by generating a new sample "from scratch" when returning to the root node and then iteratively stepping through the dimensions of the random vector.
Furthermore, \ts\ can make use of powerful neural networks for approximating conditionals of the target, thus caching computations for related inference problems.
Although, adaptive MCMC methods exist, they usually only consider small sets of adaptive parameters \cite{andrieu2008tutorial}.
Recently, MCMC methods have been extended to transition operators generated by neural networks, which are trained either by adversarial training, meta learning or mixing time criteria \cite{song2017nice, levy2017generalizing, neklyudov2018metropolis, wang2018meta}.
However, these were formulated for continuous domains and rely on differentiability  and thus do not carry over straight-forward to discrete domains.

Our proposed algorithm is closely related to Sequential Monte Carlo (SMC) methods \cite{del2006sequential}, another class of broadly applicable inference algorithms.
Often, these methods are applied to generate approximate samples by sequentially sampling the dimensions of a random vector, \eg in particle filtering for temporal inference problems \cite{doucet2009tutorial}. 
Usually, these methods do not allow for backtracking, \ie re-visiting previously discarded partial configurations, although few variants with some back-tracking heuristics do exist \cite{klepal2008backtracking, grassberger2004sequential}.
This is contrast to the \ts\ algorithm, which decides at every iteration where to expand the current tree based on a full tree-traversal from the root and therefore allows for backtracking an arbitrary number of steps.
Furthermore, we propose to train value functions which approximately marginalize over the "future" (\ie variables following the one in question in the ordering), thus taking into account relevant 
downstream effects.
\cite{gu2015neural,kempinska2017adversarial} introduce adaptive NN proposals, \ie value functions in our formulation, but these are trained to match the "filtering" distribution,
thus they do not marginalize over the future.
In the decision-making formulation, this corresponds to learning proposals based on immediate rewards instead of total returns.
However, recent work in continuous domains has begun to address this \cite{guarniero2017iterated, heng2017controlled, lawson2018twisted, piche2018probabilistic}, however, they do not make use of guided systematic search.

Recently, distilling inference computations into parametric functions as been extended to discrete distributions based on the framework of variational inference.  \cite{mnih2014neural, mnih2016variational} highlight connections to the REINFORCE gradient estimator \cite{williams1992simple} and  propose various value function-like control variates for reducing its variance.
Multiple studies propose to utilize continuous relaxation of discrete variables to make use of so-called reparametrization gradients for learning inference computations, \eg \cite{maddison2016concrete}.

In addition to the approximate inference methods discussed above, there are numerous algorithms for exact inference in discrete models.
One class of methods called Weighted Model Counting (WMC) algorithms, is based on representing the target probability distribution as Boolean formulas with associated weights, and convert inference into the problem of summing weights over satisfying assignments of the associated SAT problem \cite{chavira2008probabilistic}. 
In particular, it has been shown that DPLL-style SAT solvers \cite{davis1962machine} can be extended to exactly solve general discrete inference problems \cite{sang2005solving, bacchus2009solving}, often outperforming other standard methods such as the junction tree algorithm \cite{lauritzen1988local}.
Similar to \ts, this DPLL-based approach performs inference by search, \ie it recursively instantiates variables of the SAT problem.
Efficiency is gained by chaching solved sub-problems \cite{bacchus2003algorithms} and heuristics for adaptively choosing the search order of variables \cite{sang2005solving}.
We expect that similar techniques could be integrated into the \ts\ algorithm, potentially greatly improving its efficiency.
In contrast to WMC methods, \ts\ dynamically chooses the most promising sub-problems to spend compute on via the UCT-like selection rule which is informed by all previous search tree expansions.

\section{Discussion}

Structured distributions $P^\ast_X$, such as factor graphs, Bayesian networks etc, allow for a very compact representation of an infinitely large set of beliefs,
\eg $P^\ast_X$ implies beliefs over $f(X)$ for every test function $f$, including marginals, moments etc.
This immediately raises the question: "What does it mean to 'know'" a \emph{distribution}? (paraphrased from \cite{diaconis1988bayesian}).
Obviously, we need to perform probabilistic inference to "convert the implicit knowledge" of $P^\ast_X$ (given by \eg factors) into "explicit knowledge" in terms of the beliefs of interest (quoted from \cite{gershman}).
If the dimension of $X$ is anything but very small, this inference process cannot be assumed to be "automatic", but ranks among the most complex computational problems known, and large amounts of computational resources have to be used to just approximate the solution.
In other challenging computational problems such as optimization, integration or solving ordinary differential equations,
it has been argued that the results of computations that have not yet been executed are to be treated as unobserved variables, and knowledge about them to be expressed as beliefs \cite{movckus1975bayesian, diaconis1988bayesian}.
This would imply for the inference setting considered in this paper, that we should introduce second-order, or meta-beliefs over yet-to-be-computed first-order beliefs implied by $P^\ast_X$.
Approximate inference could then proceed analogously to Bayesian optimization: Evaluate the factors of $P^\ast_X$ at points that result in the largest reduction of second-order uncertainty over the beliefs of interest.
However, it is unclear how such meta-beliefs can be treated in a tractable way.
Instead of such a full Bayesian numerical treatment involving second-order beliefs, we adopted cheaper Upper Confidence Bound heuristics for quantifying uncertainty.

For sake of simplicity, we assumed in this paper that the computational cost of inference is dominated by evaluations of the factor oracles.
This assumption is well justified \eg in applications, where some factors represent large scale  scientific simulators \cite{baydin2019etalumis},
or in modern deep latent variable models, where a subset of factors is given by deep neural networks that take potentially high-dimensional observations as inputs.
If this assumption is violated, \ie all factors can be evaluated cheaply,
the comparison of \ts\ to SMC and other inference methods will become less favourable for the former. 
\ts\ incurs an overhead for traversing a search tree before expanding it, attempting to use the information of all previous oracle evaluations. 
If these are cheap, a less sequential and more parallel approach, such as SMC, might become more competitive.

We expect that \ts\ can be improved and extended in many ways.
Currently, the topology of the factor graph is only partially used for the reward definition and potentially for graph net value functions.
One obvious way to better leverage it would be to check if after conditioning on a prefix $x_{<n}$, corresponding to a search depth $n$, the factor graph decomposes into independent components that can be solved independently. 
Furthermore, \ts\ uses a fixed ordering of the variables. 
However, a good variable ordering can potentially make the inference problem much easier.
Leveraging existing or developing new heuristics for a problem-dependent and dynamic variable ordering could potentially increase the inference efficiency of \ts.

\clearpage
\bibliographystyle{unsrt}
\bibliography{references}

\clearpage

\clearpage
\appendix

\section{Details for \ts ~ algorithm}

We define a search tree $\tree$ in the following way.
Nodes in $\tree$ at depth $n\in\{0,1,\ldots, N\}$ are indexed by the (partial) state $x\in\{1,\ldots,K\}^n$, and the root is denoted by $\emptyset$.
Each node $x$ at depth $n=len(x)$ keeps track of the corresponding reward evaluation $R_n(x)$ and the following quantities for all its children:
\begin{enumerate}
    \item visit counts $\eta_{n+1}(\cdot\vert x)=(\eta_{n+1}(1\vert x),\ldots,\eta_{n+1}(K\vert x))\in \mathbb N^K$ over the children,
    \item state-action values $Q_{n+1}(\cdot\vert x)\in\mathbb R^K$,
    \item prior state-action values $Q^\phi_{n+1}(\cdot\vert x)\in\mathbb R^K $, and
    \item a boolean vector $C_{n+1}(\cdot\vert x)\in\{0,1\}^K$ if its children are complete (\ie fully expanded, see below).
\end{enumerate}

Standard MCTS with (P)UCT-style tree traversals applied to the inference problem can in general visit any state-action pair multiple times;
this is desirable behavior in general MDPs with stochastic rewards, where reliable reward estimates require multiple samples.
However, the reward $R$ in our MDP is deterministic as defined in \eqref{eq:reward_def},
and therefore there is no benefit in re-visiting fully-expanded sub-trees.
To prevent the \ts\ algorithm from doing so, we explicitly keep track at each node if the sub-tree rooted in it is fully-expanded; such a node is called complete.
Initially no internal node is complete, only leaf nodes at depth $N$ are tagged as complete.
In the backup stage of the tree-traversal, we tag a visited node as complete if it is a node of depth $N$ (corresponding to a completed sample) or if all its children are complete.
We modify the action selection \eqref{eq:q-uct} such that the $\argmax$ is only taken over actions not leading to complete sub-trees.
The \ts\ algorithm is given in full in \algref{alg:ts-details}.

\begin{algorithm}[t]\caption{\ts ~ procedures}\label{alg:ts-details}
\begin{algorithmic}[1]
\begin{small}
\State {\bf globals} reward function $R$, prior state-action value function $Q^\phi$
\newline
\Procedure{\ts}{budget B}
    \State initialize empty tree $\tree \leftarrow \emptyset$
    \State available budget $b\leftarrow B$
    \While{$b>M$}
        \State $\tree,\Delta b$ \ $\leftarrow$\ \textsc{TreeTravsersal} ($\tree$)
        \State $b\leftarrow b-\Delta b$
    \EndWhile
    \State {\bf return} tree $\tree$
\EndProcedure
\newline
\Procedure{TreeTraversal}{tree $\tree$}
    \Statex ~ ~ ~ // traversal
    \State $x\leftarrow \emptyset$
    \While{$x\in\tree$}
        \State $n\leftarrow len(x)$
        \State $a \leftarrow$ \textsc{Q-UCT}$(\eta_{n+1}(\cdot\vert x), Q_{n+1}(\cdot\vert x), Q_{n+1}^\phi(\cdot\vert x), C_{n+1}(\cdot\vert x))$
        \State $x \leftarrow x \circ a$            
    \EndWhile
    \Statex  ~ ~ ~ // expansion
    \If{$x\not\in\tree$}
        \State $\tree\ \leftarrow \ \tree\ \cup $ \ \textsc{Expand}$(x)$
        \State used budget $\Delta b\leftarrow \vert M_n\vert$ \ \ \ \   // see def.\ \ref{def:reward}
    \EndIf
    \Statex  ~ ~ ~ // backup
    \For{$n=len(x),\ldots,1,0$}
        \State $V_{n+1}(x_{\leq n}) = \log \sum_{a'=1}^K \exp Q_{n+1}(a'\vert x_{\leq n})$
        \State $Q_n(x_n\vert x_{<n}) \leftarrow R_n(x_n\vert x_{<n}) + V_{n+1}(x_{\leq n})$
        \State $C_n(x_n\vert x_{<n})\leftarrow \min_{a'}C_{n+1}(a'\vert x_{\leq n})$
        \State $\eta_n(x_n\vert x_{<n})\leftarrow \eta_n(x_n\vert x_{<n})+1$
    \EndFor
    \State {\bf return} $\tree$, $\Delta b$
\EndProcedure
\newline
\Procedure{Q-UCT}{$\eta_{n+1}(\cdot\vert x)$, $Q_{n+1}(\cdot\vert x)$, $Q_{n+1}^\phi(\cdot\vert x)$, $C_{n+1}(\cdot\vert x)$}
    \State {\bf return} $\argmax$ of \eqref{eq:q-uct} over in-complete children $\{a\vert C_{n+1}(a\vert x)=0\}$
\EndProcedure
\newline
\Procedure{Expand}{state $x$}
    \State $n\leftarrow len(x)$
    \State evaluate reward function $ R_n(x_n\vert x_{<n})$
    \State initialize $\eta_{n+1}(\cdot\vert x)\leftarrow(0,\ldots,0)$
    \If{$n=N$} \ \ \ // $x$ is leaf
        \State initialize $Q_{n+1}(a\cdot\vert x) \leftarrow -\log K$ for all $a\in\{1,\ldots,K\}$        
        \State initialize $C_{n+1}(\cdot\vert x)\leftarrow(1,\ldots,1)$
    \Else
        \State evaluate prior $Q_{n+1}^\phi(\cdot\vert x)$
        \State initialize $Q_{n+1}(\cdot\vert x) \leftarrow Q_{n+1}^\phi(\cdot\vert x)$        
        \State initialize $C_{n+1}(\cdot\vert x)\leftarrow(0,\ldots,0)$
    \EndIf
    \State {\bf return} node $x$ with $\eta_{n+1}(\cdot\vert x)$,
                        $Q_{n+1}(\cdot\vert x)$, 
                        $Q_{n+1}^\phi(\cdot\vert x)$,
                        $C_{n+1}(\cdot\vert x)$,
                        $ R_n(x_n\vert x_{<n})$
\EndProcedure
\end{small}
\end{algorithmic}
\end{algorithm}

\section{Proofs}
\subsection{Observation \ref{obs:q-values}}
\begin{proof}
This observation has been proven previously in the literature, but we will give a short proof here for completeness.
We show the statement by determining the optimal policy and value function by backwards dynamic programming (DP). 
We anchor the DP induction by defining the optimal value function at step $N+1$ as zero, \ie $V^\ast_{N+1}(x_{\leq N})=0$.
Using the law of iterated expectations, we can decompose the optimal value function in the following way for any $n=N,\ldots,1$:
\begin{eqnarray*}
V^\ast_{n}(x_{< n})     &=& \max_{\pi_n,\ldots,\pi_N}\  \E_{P^\pi_{X_{\geq n}\vert x_{<n}}}\left[\sum_{n'=n}^N R_{n'}-\log\pi_{n'}\right]\\
                        &=& \max_{\pi_n}\ \E_{P^\pi_{X_{n}\vert x_{<n}}}\ \max_{\pi_{n+1},\ldots,\pi_N}\ \E_{P^\pi_{X_{> n}\vert x_{<n}\circ X_n}}\left[\sum_{n'=n}^N R_{n'}-\log\pi_{n'} \right]\\
                        &=& \max_{\pi_n}\ \E_{P^\pi_{X_{n}\vert x_{<n}}} \left[R_{n}-\log\pi_{n} 
                        + \max_{\pi_{n+1},\ldots,\pi_N}\E_{P^\pi_{X_{> n}\vert x_{<n}\circ X_n}}\left[\sum_{n'=n+1}^N R_{n'}-\log\pi_{n'}  \right]
                        \right]\\
                        &=& \max_{\pi_n}\ \E_{P^\pi_{X_{n}\vert x_{<n}}}\ \left[R_n-\log\pi_n +V^\ast_{n+1}\right]
                        .   
\end{eqnarray*}
Therefore, assuming by induction that $V^\ast_{n+1}$ has been computed, we can find the optimal policy $\pi^\ast_n$ and value $V^\ast_n$ at step $n$ by solving:
\begin{subequations}
\begin{alignat}{2}
&\argmax_{f:\{1,\ldots,K\}\rightarrow [0,1]}        &\qquad &\sum_{a=1}^K f(a)\cdot \left(R_n(x_{<n}\circ a)-\log f(a)+V^\ast_{n+1}(x_{<n}\circ a)\right)\label{eq:pi_obj}\\
&\mbox{subject to}  &       & \sum_{a=1}^K f(a)=1.%\nonumber 
%\\ &                   &       & f(a)\geq 0, \ \   \forall a\in\{1,\ldots, K\}\nonumber .
\end{alignat}
\end{subequations}
The solution to this optimization problem can be found by the calculus of variations (omitted here) and is given by:
\begin{eqnarray*}
\log \pi^\ast(x_n\vert x_{<n})  &\propto&  R_n(x_{\leq n}) + V_{n+1}^\ast(x_{\leq n}) = Q^\ast_n(x_n\vert x_{<n}),
\end{eqnarray*}
where we used the definition of the optimal state-action value function.
Furthermore, at the optimum, the objective \eqref{eq:pi_obj} assumes the value:
\begin{eqnarray*}
V^\ast_{n}(x_{< n}) &=& \log\sum_{x_n=1}^K\exp Q_n^\ast(x_n\vert x_{<n}).
\end{eqnarray*}
This expression, together with the definition of $Q^\ast$ establishes the soft-Bellman equation.
The optimal value $V^\ast_{n+1}$ is also exactly the log-normalizer for $\pi^\ast_n$. Therefore, we can write:
\begin{eqnarray*}
\log\pi^\ast_n(x_n\vert x_{<n}) = Q^\ast_n(x_n\vert x_{<n})- V^\ast_{n+1}(x_{\leq n}).
\end{eqnarray*}
\end{proof}

\subsection{Proof of Lemma \ref{lem:exact_values}}
\begin{figure}
\centering
\begin{tikzpicture}[scale=1.0, >=stealth]
\tikzstyle{empty}=[]
\tikzstyle{incomp}=[circle, inner sep=1pt, minimum size = 2.5mm, thick, draw =black!80, node distance = 5mm, scale=0.75]
\tikzstyle{comp}=[circle, fill =black!80, inner sep=1pt, minimum size = 2.5mm, thick, draw =black!80, node distance = 5mm, scale=0.75]
\tikzstyle{directed}=[->, thick, shorten >=0.5 pt, shorten <=1 pt]
\tikzstyle{onpath}=[->, thick, shorten >=0.5 pt, shorten <=1 pt, blue]

\node[incomp] at (0,0) (root) [label=above:$x_{\leq n}$] {};
\node[incomp] at (-1,-1.5) (l10) [] {};
\node[comp] at (0,-1.5) (l11) [label=below:$\stackrel{\swarrow\downarrow\searrow}{\ldots}$] {};
\node[comp] at (1,-1.5) (l12) [label=below:$\stackrel{\swarrow\downarrow\searrow}{\ldots}$] {};
\node[comp] at (-2,-3) (l20) [label=below:$\stackrel{\swarrow\downarrow\searrow}{\ldots}$] {};
\node[comp] at (-1,-3) (l21) [label=below:$\stackrel{\swarrow\downarrow\searrow}{\ldots}$] {};
\node[incomp] at (0,-3) (l22) [label=right:$x_{\leq m}$] {};
\node[comp] at (-1,-4.5) (l30) [] {};
\node[incomp] at (0,-4.5) (l31) [label=below:{leaf $x_{\leq N}$}] {};
\node[comp] at (1,-4.5) (l32) [] {};
\path   (root) edge [onpath] (l10)
        (root) edge [directed] (l11)
        (root) edge [directed] (l12)
        (l10) edge [onpath] (l22)
        (l10) edge [directed] (l20)
        (l10) edge [directed] (l21)
        (l22) edge [directed] (l30)
        (l22) edge [onpath] (l31)
        (l22) edge [directed] (l32)
;
\end{tikzpicture}
\caption{{\bf Completing a sub-tree yields the exact $Q$-value.} Assume the tree traversal shown in blue completes the sub-tree $\tree_{x_{\leq n}}$ rooted in $x_{\leq n}$. Then, by construction, the soft-Bellmann backups along this path, at every intermediate node $x_{\leq m}$ for $m>n$, take as input the values of all children. By construction, all but one children correspond to complete sub-trees of a smaller depth; these have the correct values by induction.
The other remaining child corresponds to a sub-tree that was completed by the last traversal and therefore has also the correct value.}
\label{fig:tree_completion}
\end{figure}
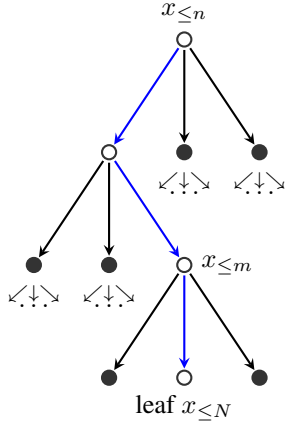

\begin{proof}
We will show this statement by induction on the depth $d:=N-n$ of the sub-tree $\tree_{x_{\leq n}}$ with root $x_{\leq n}$.
For $d=0$, \ie $n=N$, the state-action values $Q_{N+1}(\cdot\vert x_{\leq N}):=-\log K$ are defined such that $V_{N+1}(x_{\leq N})=0$, which is the correct value.
Consider now the general case $1\leq d \leq N$. 
Let $\tree'_{x_{\leq n}}$ be the sub-tree \emph{before} the last tree traversal that expanded the last 
missing node $x_{\leq N}$, ie $\tree_{x_{\leq n}}=\tree'_{x_{\leq n}}\cup x_{\leq N}$;
for an illustration see \figref{fig:tree_completion}.
The soft-Bellman backups of the last completing tree-traversal on the path leading to $x_{\leq N}$ are by construction all of 
the following form: For any node $x_{\leq m}$ on the path, all children except for one correspond to already completed sub-trees (before the last traversal). The sub-tree of the one remaining child is completed by the last traversal. 
All complete sub-trees on the backup path are of depth smaller than $d$ and therefore by induction their roots have the correct values $V_{m+1}^\ast(x_{\leq m})$. 
Hence evaluating the soft-Bellman backup \eqref{eq:soft-B} (with the true noiseless reward $R$) 
yields the correct value for $x_{\leq n}$.
\end{proof}

\section{Details for Experiments}

\subsection{Baseline Inference Methods}
\subsubsection{SIS and SMC}
For each experiment we determined the number of SIS and SMC particles $I$ such that the entire budget $B$ was used.
We implemented SMC with an resampling threshold $t\in[0,1]$, \ie a resampling step was executed when the effective sample size (ESS) was smaller than $tI$.
The threshold $t$ was treated as a hyperparameter; SMC with $t=0$ was used as SIS results.

\subsubsection{BP}
We used the algorithm outline on p.\ 301 from \cite{mezard2009information}.
For generating a single approximate sample from the target, the following procedure was executed.
Messages from variable to factor nodes were initialized as uniform; 
then $N_{message}$ message-passing steps, each consisting of updating factor-variable and variable-factor messages were performed.
$X_1$ was then sampled form the resulting approximate marginal, and the messages from $X_1$ to its neighboring factors were set to the corresponding atom.
This was repeated until all variables $X_n$ were sampled, generating one approximate sample from the joint $P^\ast_X$.

In total, we generated multiple samples with the above algorithm such that the budget $B$ was exhausted. 
The number $N_{message}$ of message-passing steps before sampling each variable $X_n\vert X_{<n}$ was treated as a hyperparameter.

\subsubsection{GIBBS}
We implemented standard Gibbs sampling. All variables were initially drawn uniformly from $\{1,\ldots, K\}$ , and $N_{Gibbs}$ iterations, each consisting of updating all variables in the fixed order $X_1$ to $X_N$, were executed.
This generated a single approximate sample.
We repeated this procedure to generate multiple samples such that the budget $B$ was exhausted. We treated $N_{Gibbs}$ as a hyperparameter.

\subsection{Hyperparameter optimization}

For each inference method (except for SIS) we optimized one hyperparameter on a initial set of experiments. 
For \ts, we fixed $\epsilon=0.1$ and optimized $c$ from \eqref{eq:q-uct}.
Different hyperparameter values were used for different families of distributions.
Hyperparameters were chosen such as to yield lowest $\Delta\KL$.

\subsection{Details for Synthetic Distributions}
\subsubsection{Chains}
The unary potentials $\psi_n(x_n)$ for $n=1,\ldots,N$ for the chain factor graphs where randomly generated in the following way.
The values of $\psi_n(x_n=k)$ for $n=1,\ldots,N$ and $k=1,\ldots,K$ where jointly drawn from a GP over the two dimensional domain $\{1,\ldots,N\}\times\{1,\ldots,K\}$ with an RBF kernel with bandwidth 1 and scale 0.5.
Binary potentials $\psi_{n,n+1}(x_n,x_{n+1})$ were set to $2.5\cdot d(x_n,x_{n+1})$, where $d(x_n,x_{n+1})$ is the distance between $x_n$ and $x_{n+1}$ on the 1-d torus generated by constraining $1$ and $K$ to be neighbors.

\subsubsection{PermutedChains}

We first uniformly drew random permutations $\sigma:\{1,\ldots,N\}\rightarrow\{1,\ldots,N\}$.
We then randomly generated conditional probability tables for $P^\ast_{X_{\sigma(n)}\vert x_{\sigma({n-1})}}$ by draws from a symmetric Dirichlet with concentration  parameter $\alpha=1$.
These were then used as binary factors.

\subsubsection{FactorGraphs1}
We generated factor graphs for this family in the following way. First, we constructed Erd\H os-R\'enyi random graphs with $N$ nodes with edge probability $p=2\log(N)/N$; graphs with more than one connected component were rejected.
For each clique in this graph we inserted a random factor and connected it to all nodes in the clique; graphs with cliques of size $>4$ where rejected.

For applying the sequential inference algorithms \ts, SIS and SMC, variables in the graph were ordered by a simple heuristic.
While iterating over factors in order of descending degree, all variables in the current factor were were added to the ordering until all were accounted for.

\subsubsection{FactorGraphs2}
We generated factor graphs for this family over binary random variables $K=2$ in the following way.
Variables $X_{2n+1}$ and $X_{2(n+1)}$ for $n=0,\ldots, N/2-1$ were connected with a NOT factor, which carries out the computation $ XOR(X_{2n+1},X_{2(n+1)})$.
We then constructed Erd\H os-R\'enyi random graphs of size $N/2$ over all pairs of nodes $(X_{2n+1},X_{2(n+1)})$ with edge probability $p=3\log(N/2)/N$; graphs with more than one connected component were rejected.
For each clique in this intermediate graph we inserted a MAJORITY factor and connected it to either to $X_{2n+1}$ or $X_{2(n+1)}$; graphs with cliques of size $>4$ where rejected.
MAJORITY factors return a value of 1.0 if half or more nodes in its neighborhood are 2 and return 0 otherwise.
The output values of all factors were also scaled by 2.0; otherwise the
resulting distributions were found to be very close to uniform.

\subsection{Details For Experiments w/ Value Functions}
All neural networks were trained with the ADAM optimizer \cite{kingma2014adam} with a learning rate of $3\cdot 10^{-4}$ and mini-batches of size 128.
The replay buffer size was set to $10^4$.

The MLP value function used for the experiment consisted of 4 hidden layers with 256 units each with RELU activation functions.
Increasing the number of units or layers did not improve results.

The GNN value function $Q^\phi(\cdot\vert x_{\leq n})$ was designed as follows.
Each variable node in the factor graph was given a $(16+K)$-dimensional feature vector, and each edge node a $16$-dimensional feature vector.
The input prefix $x_{\leq n}$ was encoded in a one-hot manner in the first $K$ components of the variable feature vectors for variables up to $n$;
for variables $>n$ the first $K$ components were set to 0. All edge features were initialized to 0.
All node and edge block networks where chosen to be MLPs with 3 hidden layers and 16 units each with RELU activations. 
Results did not improve with deeper or wider networks.
The resulting GNN was iterated 4 times; interestingly more iterations actually reduced final performance somewhat.
The output feature vector of the GNN at variable node $x_n$ was then passed to a linear layer with $K$ outputs yielding the vector $(Q^\phi(x_{n+1}=1\vert x_{\leq n}),\dots,Q^\phi(x_{n+1}=K\vert x_{\leq n}))$.

\begin{figure}[t]
    \centering
    \includegraphics[width=0.49\textwidth]{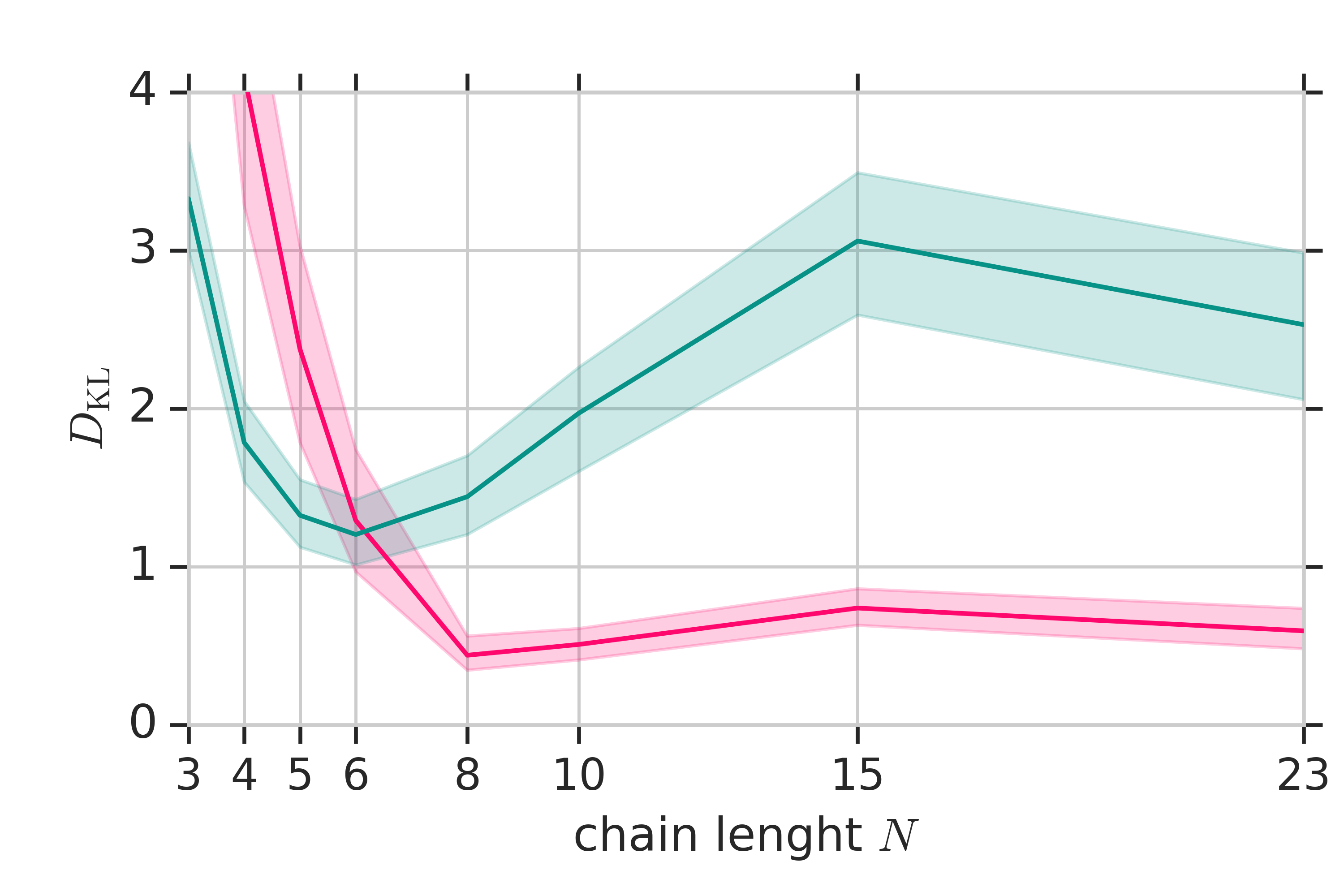}
    \caption{Approximation error for inference in chain graphs as a function of varying chain length $N$; the number $K$ of states per variable was abjusted such that 
    the total domain size $N\log K$ stayed roughly constant. \ts\ (red) performed worse than SMC (turquoise) for short and wide chains, but performs better everywhere else.}
    \label{fig:vary_depth}
\end{figure}

\end{document}